\documentclass[11pt,a4paper]{article}
\usepackage[hyperref]{acl2018}
\usepackage{times}
\usepackage{latexsym}
\usepackage{graphicx}
\usepackage{url}
\usepackage{multirow}
\usepackage{tablefootnote}
\usepackage{amssymb}
\usepackage{subcaption}
\usepackage{wrapfig}
\aclfinalcopy 
\def\aclpaperid{253} 

\title{\textit{Eyes are the Windows to the Soul}: Predicting the Rating of Text Quality Using Gaze Behaviour}

\author{$^{\bigstar}$Sandeep Mathias, $^{\bigstar,\clubsuit,\diamondsuit}$Diptesh Kanojia, $^{\bigstar}$Kevin Patel, $^{\bigstar}$Samarth Agrawal\\\textbf{$^{\spadesuit}$Abhijit Mishra, $^{\bigstar}$Pushpak Bhattacharyya}\\
  $^{\bigstar}$CSE Department, IIT Bombay\\
  $^{\clubsuit}$IITB-Monash Research Academy\\
  $^{\diamondsuit}$Monash University, Australia\\
  $^{\spadesuit}$IBM Research, India\\
  {\tt $^{\bigstar,\clubsuit}$\{sam,diptesh,kevin.patel,samartha,pb\}@cse.iitb.ac.in}\\
  {\tt $^{\spadesuit}$abhijimi@in.ibm.com}
\\}

\date{}

\begin{document}
\maketitle
\begin{abstract}
Predicting a reader's rating of text quality is a challenging task that involves estimating different subjective aspects of the text, like structure, clarity, \textit{etc.} Such subjective aspects are better handled using cognitive information. One such source of cognitive information is gaze behaviour. In this paper, we show that gaze behaviour does indeed help in effectively predicting the rating of text quality. To do this, we first model text quality as a function of three properties - organization, coherence and cohesion. Then, we demonstrate how capturing gaze behaviour helps in predicting each of these properties, and hence the overall quality, by reporting improvements obtained by adding gaze features to traditional textual features for score prediction. We also hypothesize that if a reader has fully understood the text, the corresponding gaze behaviour would give a better indication of the assigned rating, as opposed to partial understanding. Our experiments validate this hypothesis by showing greater agreement between the given rating and the predicted rating when the reader has a full understanding of the text.
\end{abstract}

\section{Introduction}
\label{Introduction Section}
Automatically rating the quality of a text is an interesting challenge in NLP. It has been studied since Page's seminal work on automatic essay grading in the mid-1960s \cite{page1966imminence}. This is due to the dependence of quality on different aspects such as the overall structure of the text, clarity, \textit{etc.} that are highly qualitative in nature, and whose scoring can vary from person to person \cite{blindtruth}. 

Scores for such qualitative aspects cannot be inferred solely from the text and would benefit from psycholinguistic information, such as gaze behaviour. Gaze based features have been used for co-reference resolution \cite{ross2016leveraging}, sentiment analysis \cite{joshi2014measuring} and translation annotation complexity estimation \cite{mishra-bhattacharyya-carl:2013:Short}.
They could also be very useful for education applications, like evaluating readability \cite{mishra2017scanpath} and in automatic essay grading.

In this paper, we consider the following qualitative properties of text: Organization, Coherence and Cohesion. A text is \textbf{well-organized} if it begins with an introduction, has a body and ends with a conclusion. One of the other aspects of organization is the fact that it takes into account how the content of the text is split into paragraphs, with each paragraph denoting a single \textit{idea}. If the text is too long, and not split into paragraphs, one could consider the text to be badly organized\footnote{Refer supplementary material for example. We have placed it there due to space constraints.}.


\begin{table*}[tpb]
\centering
\resizebox{\textwidth}{!}{%
{\small
\begin{tabular}{|p{0.5\textwidth}|p{0.5\textwidth}|}
\hline
\textbf{Example} & \textbf{Comments} \\ \hline
My favourite colour is blue. I like it because it is calming and it relaxes me. I often go outside in the summer and lie on the grass and look into the clear sky when I am stressed. For this reason, I'd have to say my favourite colour is blue. & Coherent and cohesive. \\ \hline
My favourite colour is blue. I'm calm and relaxed. In the summer I lie on the grass and look up. & Coherent but not cohesive. There is no link between the sentences. However, the text makes sense due to a lot of implicit clues (blue, favourite, relaxing, look up (and see the blue sky)). \\ \hline
My favourite colour is \textit{blue}.  \textit{Blue} sports cars go \textbf{very fast}.  Driving in \textbf{this way} is dangerous and can cause many \textit{car crashes}.  I had a \textit{car accident} once and \textbf{broke my leg}.  I was very sad because I had to miss a holiday in Europe because of \textbf{the injury}. & Cohesive but not coherent. The sentences are linked by words (that are in \textit{italics} or in \textbf{bold}) between adjacent sentences. As we can see, every pair of adjacent sentences are connected by words / phrases, but the text does not make sense, since it first starts with blue, and describes missing a holiday due to injury. \\ \hline
\end{tabular}
}
}
\caption{Examples of coherence and cohesion\protect\footnotemark.}
\label{Coherence and cohesion table}
\end{table*}
A text is \textbf{coherent} if it makes sense to a reader. A text is \textbf{cohesive} if it is well connected. Coherence and cohesion are two qualities that are closely related. A piece of text that is well-connected usually makes sense. Conversely, a piece of text that makes sense is usually well-connected. However, it is possible for texts to be coherent but lack cohesion. Table \ref{Coherence and cohesion table} provides some examples for texts that are coherent and cohesive, as well as those that lack one of those qualities.






There are different ways to model coherence and cohesion. Since coherence is a measure of how much sense the text makes, it is a semantic property of the text. It requires sentences within the text to be interpreted, by themselves, as well as with other sentences in the text \cite{van1980text}.

On the other hand, cohesion makes use of linguistic cues, such as references (demonstratives, pronouns, \textit{etc.}), ellipsis (leaving out implicit words - Eg. Sam can type and I can [\textit{type}] too), substitution (use of a word or phrase to replace something mentioned earlier - Eg. How's the croissant? I'd like to have \textbf{one} too.), conjunction (and, but, therefore, \textit{etc.}), cohesive nouns (problem, issue, investment, \textit{etc.}) and lexis (linking different pieces of text by synonyms, hyponyms, lexical chains, \textit{etc.}) \cite{halliday2014cohesion}.
\footnotetext{We took the examples from this site for explaining coherence and cohesion: {\url{http://gordonscruton.blogspot.in/2011/08/what-is-cohesion-coherence-cambridge.html}}}

Using these properties, we model the overall text quality rating. We make use of a Likert scale \cite{likert1932technique} with a range of 1 to 4, for measuring each of these properties; the higher the score, the better is the text in terms of that property. We model the text quality rating on a scale of 1 to 10, using the three scores as input. In other words,
\begin{center}
$Quality(T) = Org(T) + Chr(T) + Chs(T) - 2$,
\end{center}
\noindent where $Quality(T)$ is the text quality rating of the text $T$. $Org(T)$, $Chr(T)$, and $Chs(T)$ correspond to the \textbf{\underline{Org}}anization, \textbf{\underline{C}}o\textbf{\underline{h}}e\textbf{\underline{r}}ence, and \textbf{\underline{C}}o\textbf{\underline{h}}e\textbf{\underline{s}}ion scores respectively, for the text $T$, that are given by a reader. We subtract 2 to scale the scores from a range of 3 - 12, to a range of 1 - 10 for quality.

Texts with poor organization and/or cohesion can force readers to regress \textit{i.e.} go to previous sentences or paragraphs. Texts with poor coherence may lead readers to fixate more on different portions of text to understand them. In other words, such gaze behaviour indirectly captures the effort needed by human readers to comprehend the text \cite{just1980theory}, which, in turn, may influence the ratings given by them. Hence, these properties seem to be a good indicators for overall quality of texts.

In this paper, we address the following question: \textbf{\textit{Can information obtained from gaze behaviour help predict reader's rating of quality of text by estimating text's organization, coherence, and cohesion?}} Our work answers that question in the affirmative. We found that using gaze features does contribute in improving the prediction of qualitative ratings of text by users.

Our work has the following contributions. Firstly, we propose \textbf{a novel way to predict readers' rating of text} by recording their eye movements as they read the texts. Secondly, we show that \textbf{if a reader has understood the text completely, their gaze behaviour is more reliable}. Thirdly, \textbf{we also release our dataset}\footnote{The dataset can be downloaded from \url{http://www.cfilt.iitb.ac.in/cognitive-nlp/}} to help in further research in using gaze features in other tasks involving predicting the quality of texts.

In this paper, we use the following terms related to eye tracking. The \textbf{interest area (IA)} is an area of the screen that is under interest. We mainly look at words as interest areas. A \textbf{fixation} takes place when the gaze is focused on a point of the screen. A \textbf{saccade} is the movement of gaze between two fixations. A \textbf{regression} is a special type of saccade in which the reader refers back to something that they had read earlier.

The rest of the paper is organized as follows. Section \ref{Motivation Section} describes the motivation behind our work. Section \ref{Related Work Section} describes related work in this field. Section \ref{Features Section} describes the different features that we used. Sections \ref{Experiment Setup Section} and \ref{Results Section} describes our experiments and results. Section \ref{Results Section} also contains analysis of our experiments. Section \ref{Conclusion Section} concludes our paper and mentions future work.

\begin{figure*}[t] 
\centering
\includegraphics[width=\textwidth]{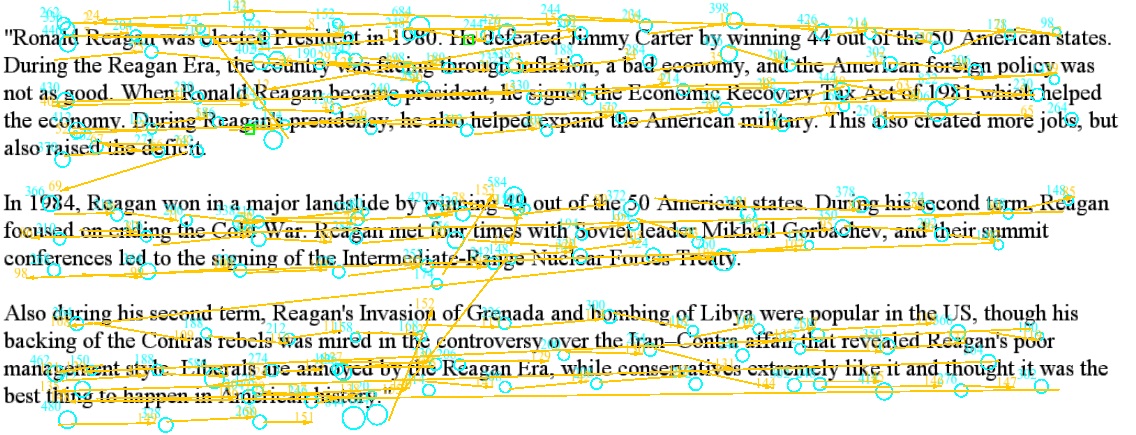}
\caption{Sample text showing fixations, saccades and regressions. This text was given scores of 4, 4, and 3 for organization, coherence and cohesion. The circles denote fixations, and the lines are saccades. Radius of the circles denote the duration of the fixation (in milliseconds), which is centred at the centre of the circle. This is the output from SR Research Data Viewer software.}
\label{Text Sample}
\end{figure*}

\section{Motivation}
\label{Motivation Section}

Reader's perception of text quality is subjective and varies from person to person. Using cognitive information from the reader can help in predicting the score he / she will assign to the text. A well-written text would not have people fixate too long on certain words, or regress a lot to understand, while a badly written text would do so. 

Figure \ref{Text Sample} shows the gaze behaviour for a sample text. The circles denote fixations, and the arrows denote saccades. If we capture the gaze behaviour, as well as see how well the reader has understood the text, we believe that we can get a clearer picture of the quality rating of the text. 

One of the major concerns is \textit{How are we going to get the gaze data?} This is because capability to gather eye-tracking data is not available to the masses. However, top mobile device manufacturers, like Samsung, have started integrating basic eye-tracking software into their smartphones (Samsung Smart Scroll) that are able to detect where the eye is fixated, and can be used in applications like scrolling through a web page. Start-ups, like Cogisen\footnote{\url{www.cogisen.com}}, have started using gaze features in their applications, such as using gaze information to improve input to image processing systems. Recently, SR Research has come up with a portable eye-tracking system\footnote{\url{https://www.sr-research.com/products/eyelink-portable-duo/}}.

\section{Related Work}
\label{Related Work Section}

A number of studies have been done showing how eye tracking can model aspects of text. Word length has been shown to be positively correlated with fixation count \cite{rayner1998eye} and fixation duration \cite{henderson1993eye}. Word predictability (i.e. how well the reader can predict the next word in a sentence) was also studied by \newcite{rayner1998eye}, where he found that unpredictable words are less likely to be skipped than predictable words.

\newcite{shermis2013handbook} gives a brief overview of how text-based features are used in multiple aspects of essay grading, including grammatical error detection, sentiment analysis, short-answer scoring, \textit{etc.} Their work also describes a number of current essay grading systems that are available in the market like \textit{E-rater}\textregistered \space \cite{attali2004automated}. In recent years, there has been a lot of work done on evaluating the holistic scores of essays, using deep learning techniques \cite{alikaniotis-yannakoudakis-rei:2016:P16-1,taghipour-ng:2016:EMNLP2016,dong-zhang:2016:EMNLP2016}.

There has been little work done to model text organization, such as \newcite{persing-davis-ng:2010:EMNLP} (using machine learning) and \newcite{taghipour2017robust} (using neural networks). However, there has been a lot of work done to model coherence and cohesion, using methods like lexical chains \cite{somasundaran-burstein-chodorow:2014:Coling}, an entity grid \cite{barzilay-lapata:2005:ACL}, \textit{etc.} An interesting piece of work to model coherence was done by \newcite{soricut-marcu:2006:POS} where they used a machine translation-based approach to model coherence. \newcite{zesch-wojatzki-scholtenakoun:2015:bea} use topical overlap to model coherence for essay grading. Discourse connectors are used as a heuristic to model cohesion by \newcite{zesch-wojatzki-scholtenakoun:2015:bea} and \newcite{persing-ng:2015:ACL-IJCNLP}. Our work is novel because it makes use of gaze behaviour to model and predict coherence and cohesion in text.

In recent years, there has been some work in using eye-tracking to evaluate certain aspects of the text, like readability \cite{gonzalezgarduno-sogaard:2017:BEA,mishra2017scanpath}, grammaticality \cite{klerke-EtAl:2015:NODALIDA}, \textit{etc.}. Our work uses eye-tracking to predict the score given by a reader to a complete piece of text (rather than just a sentence as done by \newcite{klerke-EtAl:2015:NODALIDA}) and show that the scoring is more reliable if the reader has understood the text.



\section{Features}
\label{Features Section}

In order to predict the scores of the different properties of the text, we use the following text and gaze features.

\subsection{Text-based Features}

We use a set of text-based features to come up with a baseline system to predict the scores for different properties.

The first set of features that we use are \textbf{length and count-based features}, such as word length, word count, sentence length, count of transition phrases\footnote{\url{https://writing.wisc.edu/Handbook/Transitions.html}} \textit{etc.} \cite{persing-ng:2015:ACL-IJCNLP,zesch-wojatzki-scholtenakoun:2015:bea}.

The next set of features that we use are \textbf{complexity features}, namely the degree of polysemy, coreference distance, and the Flesch Reading Ease Score (FRES) \cite{flesch1948new}. These features help in normalizing the gaze features for text complexity. These features were extracted using Stanford CoreNLP \cite{manning-EtAl:2014:P14-5}, and MorphAdorner \cite{burns2013morphadorner}.

The third set of features that we use are \textbf{stylistic features} such as the ratios of the number of adjectives, nouns, prepositions, and verbs to the number of words in the text. These features are used to model the distributions of PoS tags in good and bad texts. These were extracted using NLTK\footnote{\url{http://www.nltk.org/}} \cite{loper2002nltk}.

The fourth set of features that we use are \textbf{word embedding features}. We use the average of word vectors of each word in the essay, using Google News word vectors \cite{mikolov2013distributed}. The word embeddings are \textbf{300 dimensions}. We also calculate the mean and maximum similarities between the word vectors of the content words in adjacent sentences of the text, using GloVe word embeddings\footnote{We found that using GloVe here and Google News for the mean word vectors worked best.} \cite{pennington-socher-manning:2014:EMNLP2014}.

The fifth set of features that we use are \textbf{language modeling features}. We use the count of words that are absent in Google News word vectors and misspelled words using the PyEnchant\footnote{\url{https://pypi.python.org/pypi/pyenchant/}} library. In order to check the grammaticality of the text, we construct a 5-gram language model, using the Brown Corpus \cite{francis1979brown}.

The sixth set of features are \textbf{sequence features}. These features are particularly useful in modeling organization (sentence and paragraph sequence similarity) \cite{persing-davis-ng:2010:EMNLP}, coherence and cohesion (PoS and lemma similarity). \newcite{pitler-louis-nenkova:2010:ACL} showed that cosine similarity of adjacent sentences as one of the best predictors of linguistic quality. Hence, we also create vectors for the PoS tags and lemmas for each sentence in the text. The dimension of the vector is the number of distinct PoS tags / lemmas.

The last set of features that we look at are \textbf{entity grid features}. We define entities as the nouns in the document, and do coreference resolution to resolve pronouns. We then construct an entity grid \cite{barzilay-lapata:2005:ACL} - a 1 or 0 grid that checks whether an entity is present or not in a given sentence. We take into account sequences of entities across sentences that possess \textbf{\textit{at least}} one 1, that are either bigrams, trigrams or 4-grams. A sequence with multiple 1s denote entities that are close to each other, while sequences with a solitary 1 denote that an entity is just mentioned once and we do not come across it again for a number of sentences.

\subsection{Gaze-based Features}

The gaze-based features are dependent on the gaze behaviour of the participant with respect to interest areas.

\subsubsection*{Fixation Features}

The \textbf{First Fixation Duration} (FFD) shows the time the reader fixates on a word when he / she first encounters it. An increased FFD intuitively could mean that the word is more complex and the reader spends more time in understanding the word \cite{mishra2016predicting}.

The \textbf{Second Fixation Duration} (SFD) is the duration in which the reader fixates on a particular interest area the second time. This happens during a regression, when a reader is trying to link the word he / she just read with an earlier word.

The \textbf{Last Fixation Duration} (LFD) is the duration in which the reader fixates on a particular interest area the final time. At this point, we believe that the interest area has been processed.

The \textbf{Dwell Time} (DT) is the total time the reader fixates on a particular interest area. Like first fixation, this also measures the complexity of the word, not just by itself, but also with regard to the entire text (since it takes into account fixations when the word was regressed, \textit{etc.})

The \textbf{Fixation Count} (FC) is the number of fixations on a particular interest area. A larger fixation count could mean that the reader frequently goes back to read that particular interest area.

\subsubsection*{Regression Features}

\textbf{IsRegression} (IR) is the number of interest areas where a regression happened before reading ahead and \textbf{IsRegressionFull} (IRF) is the number of interest areas where a regression happened. The \textbf{Regression Count} (RC) is the total number of regressions. The \textbf{Regression Time} (RT) is the duration of the regressions from an interest area. These regression features could help in modeling semantic links for coherence and cohesion.

\subsubsection*{Interest Area Features}

The \textbf{Skip Count} (SC) is the number of interest areas that have been skipped. The \textbf{Run Count} (RC) is the number of interest areas that have at least one fixation. A larger run count means that more interest areas were fixated on. Badly written texts would have higher run counts (and lower skip counts), as well as fixation counts, because the reader will fixate on these texts for a longer time to understand them.

\section{Experiment Details}
\label{Experiment Setup Section}

In this section, we describe our experimental setup, creation of the dataset, evaluation metric, classifier details, \textit{etc.}

\subsection{Ordinal Classification vs. Regression}
\begin{table*}[tpb]
\centering
\resizebox{\textwidth}{!}{
\begin{tabular}{|c|c|l|} \hline
\textbf{Property} & \textbf{Grade} & \textbf{Guidelines} \\ \hline
\multirow{4}{*}{\textbf{Organization}} & 1 & \textbf{Bad}. There is no organization in the text. \\
 & 2 & \textbf{OK}. There is little / no link between the paragraphs, but they each describe an idea. \\
 & 3 & \textbf{Good}. Some paragraphs may be missing, but there is an overall link between them. \\
 & 4 & \textbf{Very Good}. All the paragraphs follow a flow from the Introduction to Conclusion. \\ \hline
\multirow{4}{*}{\textbf{Coherence}} & 1 & \textbf{Bad}. The sentences do not make sense. \\
 & 2 & \textbf{OK}. Groups of sentences may make sense together, but the text still may not make sense. \\
 & 3 & \textbf{Good}. Most of the sentences make sense. The text, overall, makes sense. \\
 & 4 & \textbf{Very Good}. The sentences and overall text make sense. \\ \hline
\multirow{4}{*}{\textbf{Cohesion}} & 1 & \textbf {Bad}. There is little / no link between any 2 adjacent sentences in the same paragraph. \\
 & 2 & \textbf{OK}. There is little / no link between adjacent paragraphs. However, each paragraph is cohesive \\
 & 3 & \textbf{Good}. All the sentences in a paragraph are linked to each other and contribute in understanding the  paragraph. \\
 & 4 & \textbf{Very Good}. The text is well connected. All the sentences are linked to each other and help in understanding the text. \\ \hline
\end{tabular}}
\caption{Annotation guidelines for different properties of text.}
\label{Guidelines Table}
\end{table*}

For each of the properties - organization, coherence and cohesion, we make use of a Likert scale, with scores of 1 to 4. Details of the scores are given in Table \ref{Guidelines Table}. For scoring the quality, we use the formula described in the Introduction. Since we used a Likert scale, we make use of ordinal classification, rather than regression. This is because each of the grades is a discrete value that can be represented as an ordinal class (where $1<2<3<4$), as compared to a continuous real number.

\subsection{Evaluation Metric}

For the predictions of our experiments, we use Cohen's Kappa with quadratic weights - quadratic weighted Kappa (QWK) \cite{cohen1968weighted} because of the following reasons. Firstly, unlike accuracy and F-Score, Cohen's Kappa takes into account whether or not agreements happen by chance. Secondly, weights (either linear or quadratic) take into account distance between the given score and the expected score, unlike accuracy and F-score where mismatches (either 1 vs. 4, or 1 vs.2) are penalized the same. Quadratic weights reward matches and penalize mismatches more than linear weights.

To measure the Inter-Annotator Agreement of our raters, we make use of Gwet's second-order agreement coefficient (Gwet's AC2) as it can handle ordinal classes, weights, missing values, and multiple raters rating the same document \cite{gwet2014handbook}.

\subsection{Creation of the Dataset}

In this subsection, we describe how we created our dataset. We describe the way we made the texts, the way they were annotated and the inter-annotator agreements for the different properties.

\subsubsection*{Details of Texts} To the best of our knowledge there isn't a publicly available dataset with gaze features for textual quality. Hence, we decided to create our own. Our dataset consists of a diverse set of \textbf{30 texts}, from Simple English Wikipedia (\textbf{10 articles}), English Wikipedia (\textbf{8 articles}), and online news articles (\textbf{12 articles})\footnote{The sources for the articles were \url{https://simple.wikipedia.org}, \url{https://en.wikipedia.org}, and \url{https://newsela.com}}. We did not wish to overburden the readers, so we kept the size of texts to approximately \textbf{200 words} each. The original articles ranged from a couple hundred words (Simple English Wikipedia) to over a thousand words (English Wikipedia). We first summarized the longer articles manually. Then, for the many articles over 200 words, we removed a few of the paragraphs and sentences. In this way, despite all the texts being published, we were able to introduce some poor quality texts into our dataset. The articles were sampled from a variety of genres, such as History, Science, Law, Entertainment, Education, Sports, \textit{etc.}

\subsubsection*{Details of Annotators}

The dataset was annotated by \textbf{20 annotators} in the \textbf{age} group of \textbf{20-25}. Out of the 20 annotators, the distribution was 9 high school graduates (current college students), 8 college graduates, and 3 annotators with a post-graduate degree.

In order to check the \textbf{eyesight} of the annotators, we had each annotator look at different parts of the screen. While they did that, we recorded how their fixations were being detected. Only if their fixations to particular parts of the screen tallied with our requests, would we let them participate in annotation.

All the participants in the experiment were \textbf{fluent speakers of English}. A few of them scored over 160 in GRE Verbal test and/or over 110 in TOEFL. Irrespective of their appearance in such exams, each annotator was made to take an English test before doing the experiments. The participants had to read a couple of passages, answer comprehension questions and score them for organization, coherence and cohesion (as either good / medium / bad). In case they either got both comprehension questions wrong, or labeled a good passage bad (or vice versa), they failed the test\footnote{25 annotators applied, but we chose only 20. 2 of the rejected annotators failed the test, while the other 3 had bad eyesight.}.

In order to help the annotators, they were given 5 \textbf{sample texts} to differentiate between good and bad organization, coherence and cohesion. Table \ref{Coherence and cohesion table} has some of those texts\footnote{The texts for good and bad organization are too long to provide in this paper. They will be uploaded in supplementary material.}.

\subsubsection*{Inter-Annotator Agreement}

Each of the properties were scored in the range of 1 to 4. In addition, we also evaluated the participant's understanding of the text by asking them a couple of questions on the text. Table \ref{Mean QWK Table for Annotators} gives the inter-annotator agreement for each of the 3 properties that they rated. The column \textbf{Full} shows the agreement only if the participant answered both the questions correct. The \textbf{Overall} column shows the agreement irrespective of the participant's comprehension of the text.
\begin{table}[tpb]
\centering
\begin{tabular}{|l|c|c|}
\hline
\textbf{Property} & \textbf{Full} & \textbf{Overall} \\ \hline
\textbf{Organization} & 0.610 & 0.519 \\
\textbf{Coherence} & 0.688 & 0.633 \\
\textbf{Cohesion} & 0.675 & 0.614 \\ \hline
\end{tabular}
\caption{Inter-Annotator Agreements (Gwet's AC2) for each of the properties.}
\label{Mean QWK Table for Annotators}
\end{table}




\subsection{System Details}

We conducted the experiment by following standard norms in eye-movement research \cite{holmqvist2011eye}. The display screen is kept \textbf{about 2 feet} from the reader, and the camera is placed midway between the reader and the screen. The reader is seated and the position of his head is fixed using a chin rest. 

Before the text is displayed, we calibrate the camera by having the participant fixate on \textbf{13 points} on the screen and validate the calibration so that the camera is able to predict the location of the eye on the screen accurately. After calibration and validation, the text is displayed on the screen in \textbf{Times New Roman} typeface with \textbf{font size 23}. The reader reads the text and while that happens, we record the reader's eye movements. The readers were allowed to take \textbf{as much time as they needed} to finish the text. Once the reader has finished, the reader moves to the next screen.

The next two screens each have a question that is based on the passage. These questions are used to verify that the reader did not just skim through the passage, but understood it as well. The questions were multiple choice, with 4 options\footnote{\textbf{Example Passage Text}: The text in Figure \ref{Text Sample}\\ \textbf{Question}: ``How many states did Ronald Reagan win in both his Presidential campaigns?''\\  \textbf{Correct Answer}: ``93'' (44+49)  }. The questions test literal comprehension (where the reader has to recall something they read), and interpretive comprehension (where the reader has to infer the answer from the text they read). After this, the reader scores the texts for organization, coherence and cohesion. The participants then take a short break (about 30 seconds to a couple of minutes) before proceeding with the next text. This is done to prevent reading fatigue over a period of time. After each break, we recalibrate the camera and validate the calibration again.

For obtaining gaze features from a participant, we collect gaze movement patterns using an SR Research Eye Link 1000 eye-tracker (monocular stabilized head mode, sampling rate 500Hz). It is able to collect all the gaze details that we require for our experiments. Reports are generated for keyboard events (message report) and gaze behaviour (interest area report) using SR Research Data Viewer software.

\subsection{Classification Details}

We also process the articles for obtaining the text features as described in Section \ref{Features Section}. Given that we want to show the utility of gaze features, we ran each of the following classifiers with 3 feature sets - only text, only gaze, and all features.

We split the data into a training - test split of sizes \textbf{70\%} and \textbf{30\%}. We used a Feed Forward Neural Network with \textbf{1 hidden layer} containing \textbf{100 neurons} \cite{bebis1994feed}\footnote{We also used other classifiers, like Naive Bayes, Logistic Regression and Random Forest. However, the neural network outperformed them.}. The size of the input vector was \textbf{361 features}. Out of these, there were \textbf{49 text features}, plus \textbf{300 dimension word embeddings features}, \textbf{11 gaze features}, and \textbf{1 class label}. The data was split using stratified sampling, to ensure that there is a similar distribution of classes in each of the training and test splits. The Feed Forward Neural Network was implemented using TensorFlow \cite{tensorflow2015-whitepaper} in Python. We ran the neural network over \textbf{10,000 epochs}, with a learning rate of \textbf{0.001} in \textbf{10 batches}. The loss function that we used was the \textbf{mean square error}.


In order to see how much the participant's understanding of the text would reflect on their scoring, we also looked at the data based on how the participant scored in the comprehension questions after they read the article. We split the articles into 2 subsets here - $Full$, denoting that the participant answered both the questions correctly, and $Partial$, denoting that they were able to answer only one of the questions correctly. The readers showed $Full$ understanding in \textbf{269 instances} and $Partial$ understanding in \textbf{261 instances}. We used the same setup here (same training - test split, stratified sampling, and feed forward neural network). We omit the remaining \textbf{70 instances} where the participant got none of the questions correct, as the participant could have scored the texts completely randomly.


\section{Results and Analysis}
\label{Results Section}

Table \ref{Overall Results Table} shows the results of our experiments using the feed forward neural network classifier. The first column is the property being evaluated. The next 3 columns denote the results for the Text, Gaze and Text+Gaze feature sets.

\begin{table}[h]
\centering
\begin{tabular}{|l|c|c|c|} \hline
\textbf{Property} & \textbf{Text} & \textbf{Gaze} & \textbf{Text+Gaze} \\ \hline
Organization & 0.237 & 0.394 & \textbf{0.563} \\
Coherence & 0.261 & 0.285 & \textbf{0.550} \\
Cohesion & 0.120 & 0.229 & \textbf{0.451} \\
Quality & 0.230 & 0.304 & \textbf{0.552} \\ \hline
\end{tabular}
\caption{QWK scores for the three feature sets on different properties.}
\label{Overall Results Table}
\end{table}

\begin{figure*}[t] 
\centering
\includegraphics[width=0.825\textwidth]{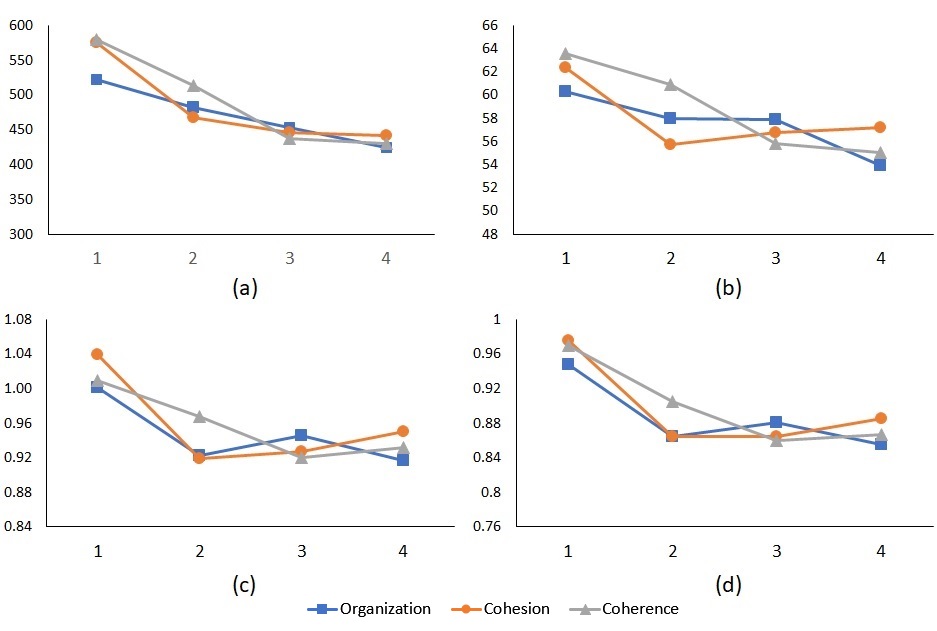}
\caption{Relation between some of the different gaze features and the score. The gaze features are (a) RD, (b) SFD, (c) FC and (d) RC. For figures (a) and (b), the units on the y-axis are milliseconds. For figures (c) and (d) the numbers are a ratio to the number of interest areas in the text. The x-axis in all 4 graphs is the score given by the annotators.}
\label{regressionChart}
\end{figure*}

The QWK scores are the predictions which we obtain with respect to the scores of all the 30 documents, scored by all 20 raters. Textual features when augmented with gaze based features show significant improvement for all the properties.

We check the statistical significance of improvement of adding gaze based features for the results in Table \ref{Overall Results Table}. To test our hypothesis - that adding gaze features make a statistically significant improvement - we run the t-test. Our null hypothesis: Gaze based features do not help in prediction, any more than text features themselves, and whatever improvements happen when gaze based features are added to the textual features, are not statistically significant. We choose a significance level of $p < 0.001$. For all the improvements, we found them to be statistically significant above this $\alpha$ level, rejecting our null hypothesis.



We also evaluate how the participant's understanding of the text affects the way they score the text. Table \ref{All Results Table} shows the results of our experiments taking the reader's comprehension into account. The first column is the property being evaluated. The second column is the level of comprehension - $Full$ for the passages where the participant answered both the questions correctly, and $Partial$ for the passages where the participant answered one question correctly. The next 3 columns show the results using the Text feature set, the Gaze feature set, and both (Text+Gaze) feature sets. From this table, we see that wherever the gaze features are used, there is far greater agreement for those with $Full$ understanding as compared to $Partial$ understanding.


\begin{table}[h]
\centering
\resizebox{\columnwidth}{!}{
\begin{tabular}{|c|c|c|c|c|}
\hline
\multicolumn{1}{|c|}{\textbf{Property}} & \textbf{Comp.} & \textbf{Text} & \textbf{Gaze} & \textbf{Text+Gaze} \\ \hline
\multirow{2}{*}{\textbf{Organization}} & Full & \textbf{0.319} & \textbf{0.319} & \textbf{0.563} \\  
 & Partial & 0.115 & 0.179 & 0.283 \\ \hline
\multirow{2}{*}{\textbf{Coherence}} & Full & 0.255 & \textbf{0.385} & \textbf{0.601} \\  
 & Partial & \textbf{0.365} & 0.343 & 0.446 \\ \hline
\multirow{2}{*}{\textbf{Cohesion}} & Full & \textbf{0.313} & \textbf{0.519} & \textbf{0.638} \\  
 & Partial & 0.161 & 0.155 & 0.230 \\ \hline
\multirow{2}{*}{\textbf{Quality}} & Full & \textbf{0.216} & \textbf{0.624} & \textbf{0.645} \\ 
 & Partial & 0.161 & 0.476 & 0.581 \\ \hline
\end{tabular}
}
\caption{QWK scores for the three feature sets on different properties categorized on the basis of reader comprehension.}
\label{All Results Table}
\end{table}

\begin{figure*}[t]
\centering
\begin{subfigure}[p]{0.3\textwidth}
\includegraphics[width=\columnwidth]{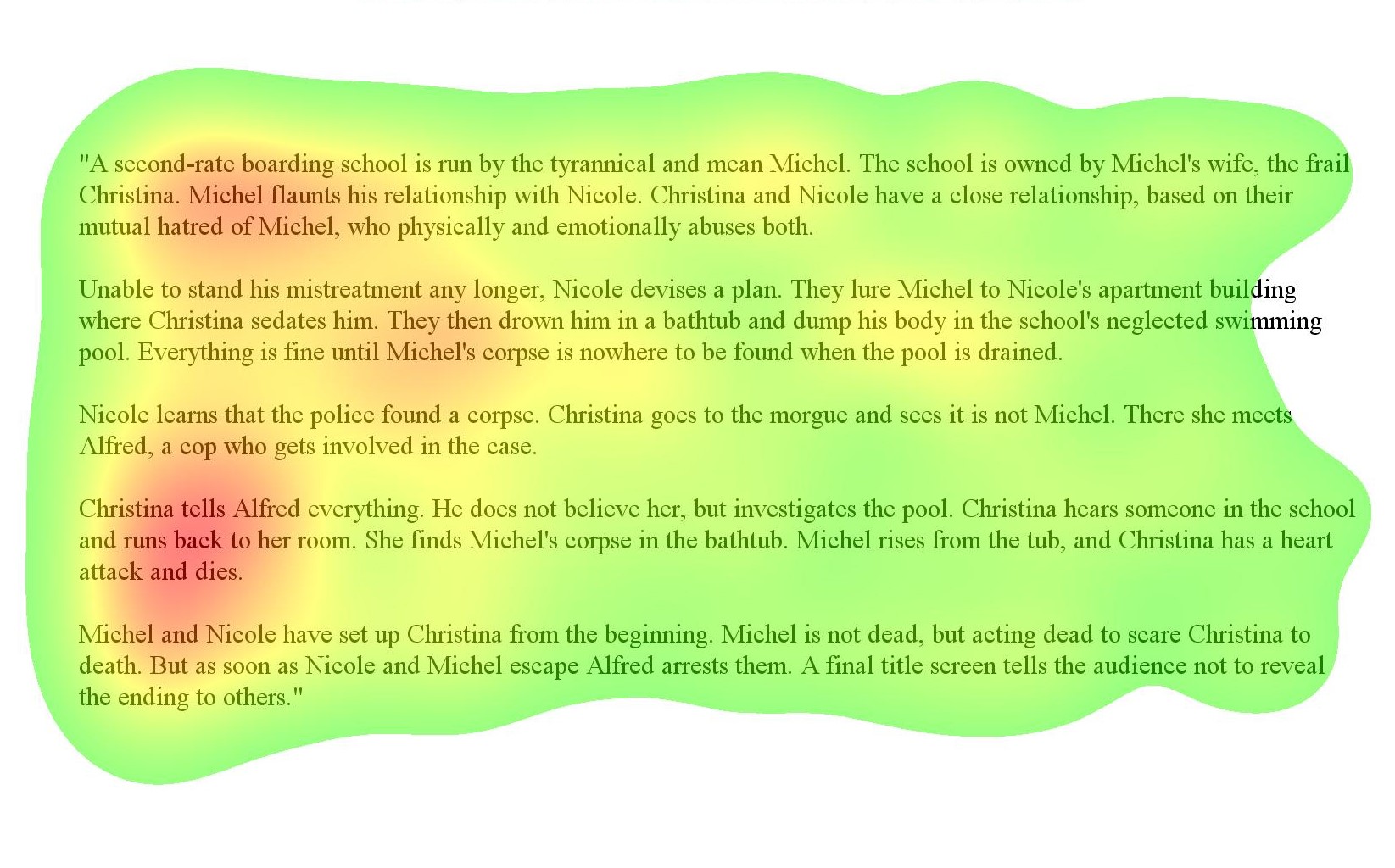}
\caption{Good (rated 10)}
\label{Good Heat Map}
\end{subfigure}
\begin{subfigure}[p]{0.3\textwidth}
\includegraphics[width=\columnwidth]{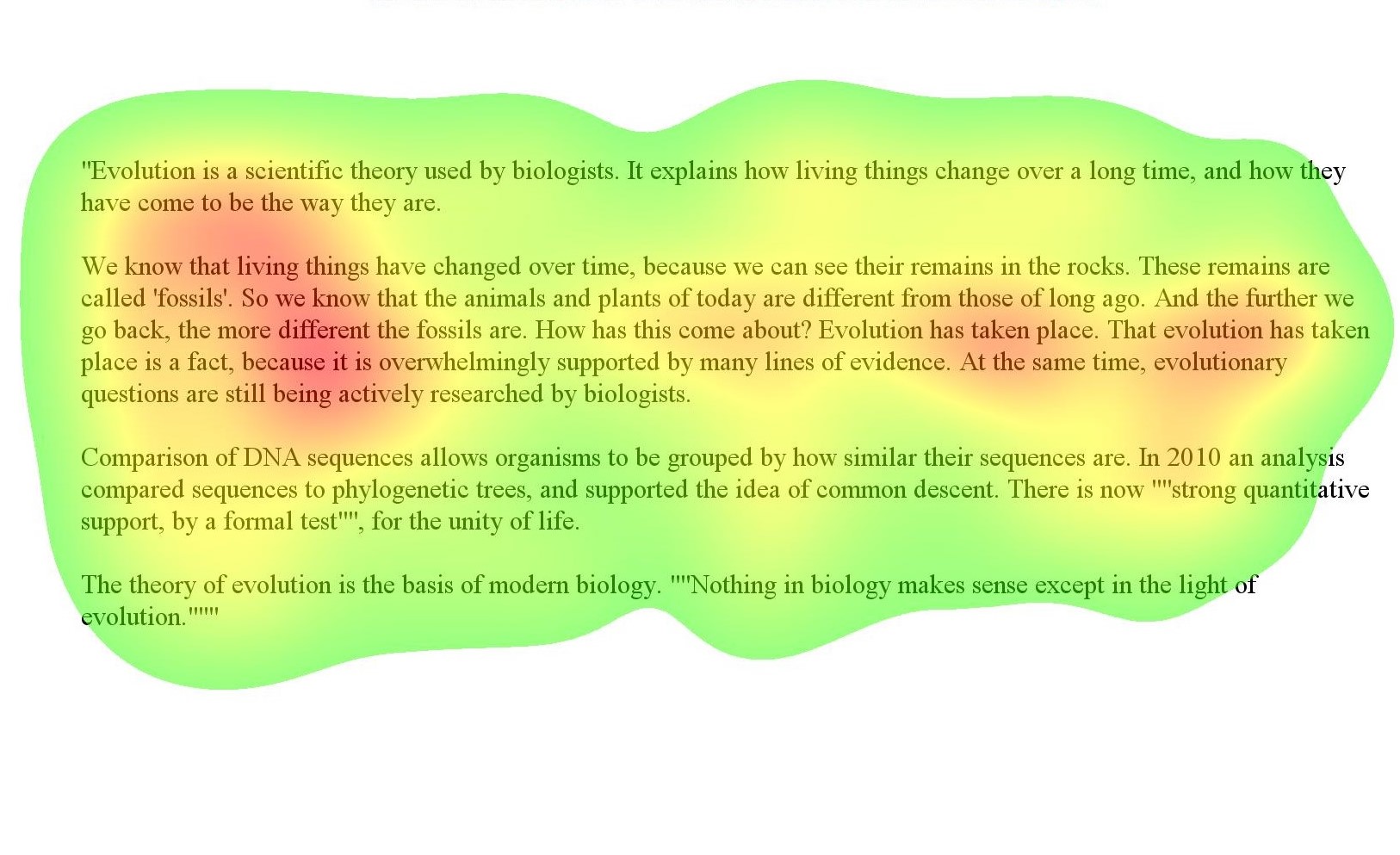}
\caption{Medium (rated 6)}
\label{Medium Heat Map}
\end{subfigure}
\begin{subfigure}[p]{0.3\textwidth}
\includegraphics[width=\columnwidth]{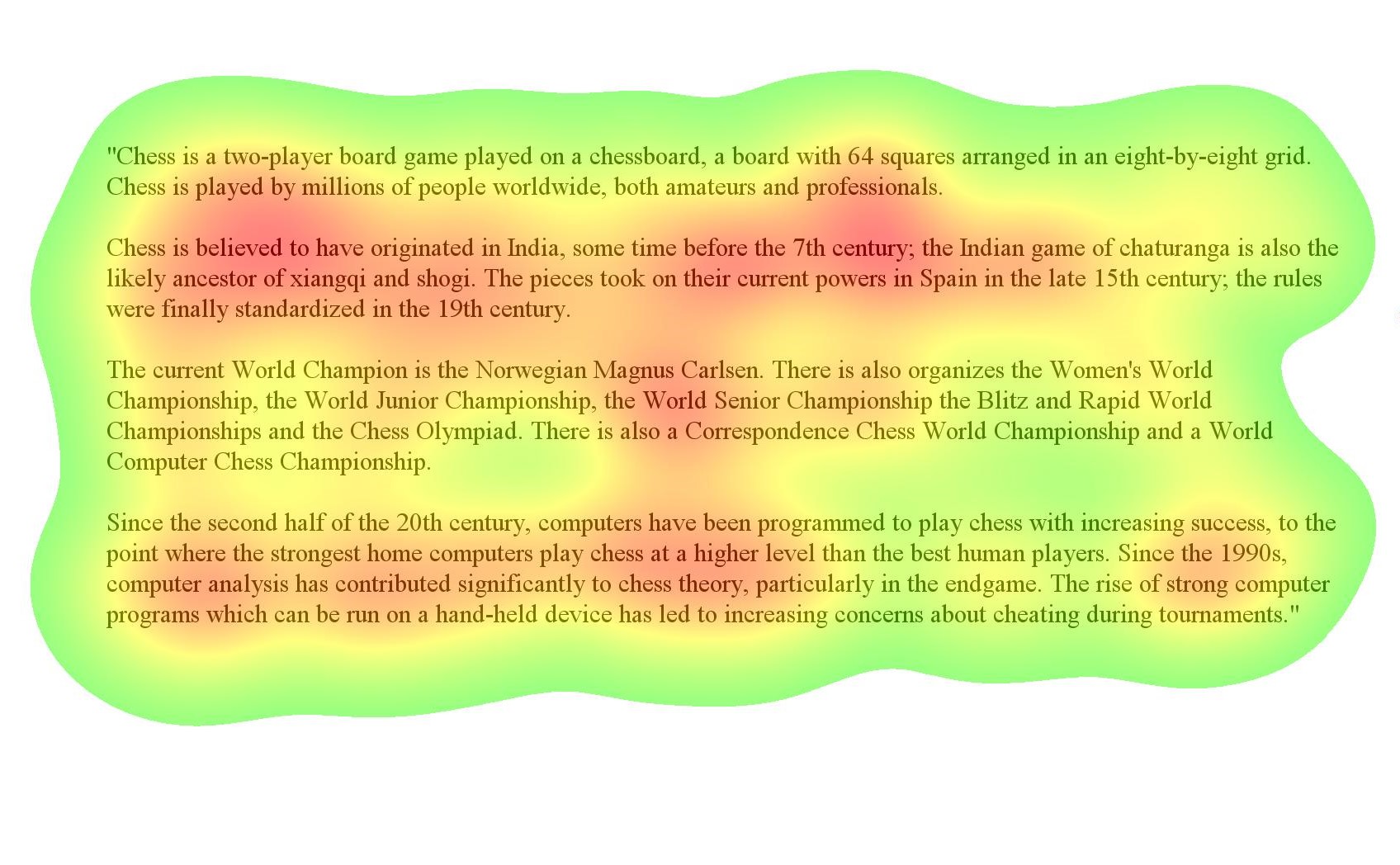}
\caption{Bad (rated 3)}
\label{Bad Heat Map}
\end{subfigure}
\caption{Fixation heatmap examples for one of the participants from SR Research Data Viewer software.}
\label{Fixation heat maps}
\end{figure*}

Figure \ref{regressionChart} shows a clear relationship between some of the gaze features and the scores given by readers for the properties - organization, cohesion and coherence. In all the charts, we see that texts with the lowest scores have the longest durations (regression / fixation) as well as counts (of fixations and interest areas fixated).

Figure \ref{Fixation heat maps} shows the fixation heat maps for 3 texts whose quality scores were good (10), medium (6) and bad (3), read by the same participant. From these heat maps, we see that the text rated good has highly dense fixations for only a part of the text, as compared to the medium and bad texts. This shows that badly written texts force the readers to fixate a lot more than well-written texts.

\subsection{Ablation Tests}

In order to see which of the gaze feature sets is important, we run a set of ablation tests. We ablate the fixations, regressions and interest area feature sets one at a time. We also ablated each of the \textbf{individual gaze features}.

\begin{table}[h]
\centering
\resizebox{\columnwidth}{!}{
{\small
\begin{tabular}{|l|c|c|c|} \hline
\textbf{Property} & \textbf{Fixation} & \textbf{Regression} & \textbf{Interest Areas} \\ \hline
Organization & -0.102 & -0.017 & \textbf{-0.103} \\
Coherence & -0.049 & -0.077 & \textbf{-0.088}  \\
Cohesion & -0.015 & \textbf{-0.040} & 0.037 \\
Quality & 0.002 & 0.016 & \textbf{-0.056} \\ \hline

\end{tabular}
}
}
\caption{\textbf{Difference} in QWK scores when ablating three gaze behaviour feature sets for different properties.}
\label{Ablation Test Table}
\end{table}

Table \ref{Ablation Test Table} gives the result of our ablation tests on the three feature sets - fixation, regression and interest area feature sets. The first column is the property that we are measuring. The next 3 columns denote the \textbf{difference} between the predicted QWK that we got from ablating the fixation, regression and interest area feature sets. We found that the Interest Area feature set was the most important, followed by fixation and regression.

Among the individual features, \textbf{Run Count} (RC) was found to be the most important for organization and quality. \textbf{First Fixation Duration} (FFD) was the most important feature for coherence, and \textbf{IsRegressionFull} (IRF) was the most important feature for cohesion. We believe that this is because the number of interest areas that are fixated on at least once and the number of interest areas that are skipped play an important role in determining how much of the text was read and how much was skipped. However, for cohesion, regression features are the most important, because they show a link between the cohesive clues (like lexis, references, \textit{etc.}) in adjacent sentences.


\section{Conclusion and Future Work}
\label{Conclusion Section}

We presented a novel approach to predict reader's rating of texts. The approach estimates the overall quality on the basis of three properties - organization, coherence and cohesion. Although well defined, predicting the score of these properties for a text is quite challenging. It has been established that cognitive information such as gaze behaviour can help in such subjective tasks \cite{mishra-bhattacharyya-carl:2013:Short,mishra2016predicting}. We hypothesized that gaze behavior will assist in predicting the scores of text quality. To evaluate this hypothesis, we collected gaze behaviour data and evaluated the predictions using only the text-based features. When we took gaze behaviour into account, we were able to significantly improve our predictions of organization, coherence, cohesion and quality. We found out that, in all cases, there was an improvement in the agreement scores when the participant who rated the text showed full understanding, as compared to partial understanding, using only the Gaze features and the Text+Gaze features. This indicated that gaze behaviour is more reliable when the reader has understood the text. 

To the best of our knowledge, our work is pioneering in using gaze information for predicting text quality rating. In future, we plan to use use approaches, like multi-task learning \cite{mishra2018cognition}, in estimating gaze features and using those estimated features for text quality prediction.


\section*{Acknowledgements}

We'd like to thank all the anonymous reviewers for their constructive feedback in helping us improve our paper. We'd also like to thank Anoop Kunchukuttan, a research scholar from the Centre for Indian Language Technology, IIT Bombay for his valuable input.

\bibliography{acl2018}
\bibliographystyle{acl_natbib}







\end{document}